\documentclass[letterpaper, 10 pt, conference]{ieeeconf}  

\IEEEoverridecommandlockouts                              

\usepackage[font=small,labelfont=bf]{caption} 
\usepackage[compress]{cite}

\makeatletter 
\newcommand{\linebreakand}{%
  \end{@IEEEauthorhalign}
  \hfill\mbox{}\par
  \mbox{}\hfill\begin{@IEEEauthorhalign}
}
\makeatother
\title{\LARGE \bf
CaRaFFusion: Improving 2D Semantic Segmentation with Camera-Radar Point Cloud Fusion and Zero-Shot Image Inpainting
}

\author{Huawei Sun$^{\star}$$^{1,2}$, Bora Kunter Sahin$^{\star}$$^{1,3}$, Georg Stettinger$^{1}$, Maximilian Bernhard$^{3}$,\\ Matthias Schubert$^{3}$, Robert Wille$^{2}$
}

\usepackage[flushleft]{threeparttable}
\usepackage{hyperref}
\usepackage{amssymb}
\usepackage{amsmath}
\usepackage{graphicx}
\usepackage{algorithm}
\usepackage{algpseudocode}
\usepackage{placeins}
\usepackage{array}
\usepackage{listings}
\usepackage{xcolor}
\usepackage{comment}
\usepackage{float}
\usepackage{subcaption}
\usepackage{multicol}
\usepackage{dblfloatfix}
\usepackage{cuted}
\usepackage{amsthm}

\begin{document}
\twocolumn[{%
\renewcommand\twocolumn[1][]{#1}%
\maketitle
\begin{center}
  \centering
    \vspace{-4mm}
  \captionsetup{type=figure}
    \includegraphics[width=0.85\textwidth]{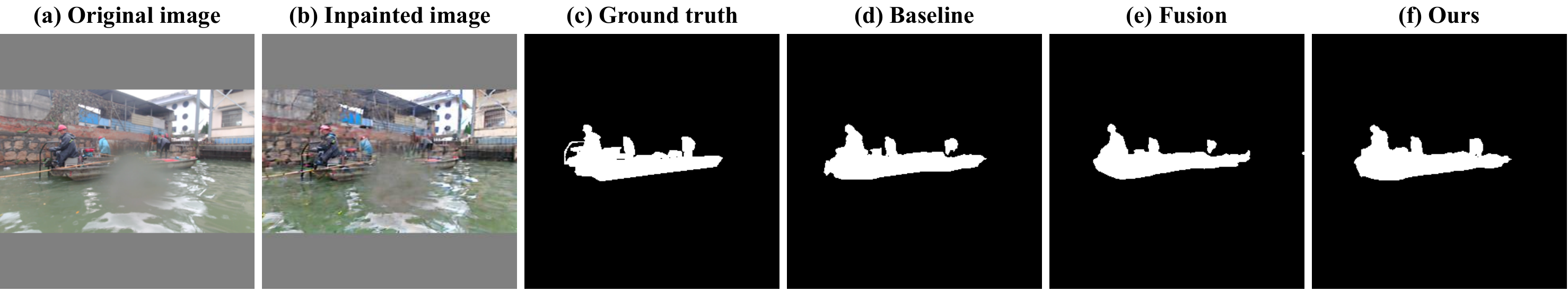}
  \captionof{figure}{
    Our method can segment the objects in very adverse conditions where other methods fail.
  }
  
  \label{fig:intro_pic}
\end{center}%
}]

\begin{abstract}
Segmenting objects in an environment is a crucial task for autonomous driving and robotics, as it enables a better understanding of the surroundings of each agent. Although camera sensors provide rich visual details, they are vulnerable to adverse weather conditions. In contrast, radar sensors remain robust under such conditions, but often produce sparse and noisy data. Therefore, a promising approach is to fuse information from both sensors.
In this work, we propose a novel framework to enhance camera-only baselines by integrating a diffusion model into a camera-radar fusion architecture. We leverage radar point features to create pseudo-masks using the Segment-Anything model, treating the projected radar points as point prompts. Additionally, we propose a noise reduction unit to denoise these pseudo-masks, which are further used to generate inpainted images that complete the missing information in the original images. Our method improves the camera-only segmentation baseline by $2.63$\% in mIoU and enhances our camera-radar fusion architecture by $1.48$\% in mIoU on the Waterscenes dataset. This demonstrates the effectiveness of our approach for semantic segmentation using camera-radar fusion under adverse weather conditions. 
\end{abstract}


\vspace{-1mm}
\def\thefootnote{1}\footnotetext{Infineon Technologies AG, Neubiberg, Germany,\\
\indent\{huawei.sun, georg.stettinger\}@infineon.com}
\def\thefootnote{2}\footnotetext{Technical University of Munich, Munich, Germany}
\def\thefootnote{3}\footnotetext{Ludwig Maximilian University of Munich, Munich, Germany, \\  \indent\{b.sahin\}@campus.lmu.de}
\def\thefootnote{*}\footnotetext{These authors contributed equally to this work}

\section{Introduction}
\label{sec:intro}

Robust semantic segmentation is vital for applications like autonomous driving~\cite{segformer}, robotics~\cite{robotics1}, and medical imaging~\cite{medical1}. Although camera-based models excel in ideal conditions, their performance deteriorates under adverse weather, leading to inconsistent boundaries and incomplete segmentations~\cite{yao2023radar}. Multi-modal integration, particularly radar-camera fusion, has emerged as a promising solution to enhance segmentation robustness~\cite{pfeuffer2019robust}.

Radars are resilient to challenging conditions, making them ideal for tasks like object detection~\cite{ob1,ob2}, depth estimation~\cite{depth1,depth2}, and BEV segmentation~\cite{radsegnet,schramm2024bevcar}. However, radar data is sparse and lacks fine-grained details, complicating its use for image-plane segmentation. Integrating radar’s spatial information with RGB’s rich visual detail poses challenges, including differences in data structure, resolution, and noise from reflections on surfaces like water.

Thus, this paper proposes CaRaFFusion, a three-stage framework for robust radar-camera fusion-based segmentation. In the first stage, spatial and visual features from the radar and camera inputs are fused via cross-attention~\cite{vaswani2017attention} to generate initial segmentation masks. The second stage uses MobileSAM~\cite{zhang2023mobilesam} to refine these masks with radar point prompts, improving robustness to weather-induced noise. To handle radar’s inherent noise, we introduce a Noise Reduction Unit (NRU) that filters unwanted reflections, producing cleaner masks. Finally, a diffusion model~\cite{rombach2022high} inpaints missing information, enhancing segmentation performance. Dual Segformer encoders~\cite{segformer} process the inpainted images and original inputs to deliver high-quality segmentation outputs.

In summary, our key contributions are as follows:

\begin{itemize}
\item We propose CaRaFFusion, a three-stage radar-camera fusion framework designed for robust segmentation under adverse conditions.
\item A Noise Reduction Unit (NRU) is invented to filter radar noise and improve segmentation mask accuracy.
\item We utilize the Image Inpainting strategy to restore information lost in adverse weather, further enhancing segmentation performance.
\end{itemize}

We train and evaluate CaRaFFusion on the WaterScenes dataset~\cite{yao2024waterscenes}, which includes radar point cloud data for 2D semantic segmentation. Results show that our approach outperforms both camera-only and existing fusion-based methods, particularly in challenging scenarios.
\section{Related Work}
\label{sec:related_work}

This section reviews semantic segmentation algorithms across image-only, radar-only, and image-radar fusion methods, and discusses generative approaches for segmentation.

\subsection{Semantic Segmentation}
We categorize semantic segmentation techniques into image-only, radar-only, and fusion-based methods.

\subsubsection{Image-Based Methods}
Image-based semantic segmentation assigns classes to image pixels, with boundary information playing a critical role in accuracy. Early approaches like Fully Convolutional Networks (FCNs)\cite{fcn} laid the foundation for CNN-based segmentation. Subsequent improved segmentation accuracy through advanced decoder modules~\cite{segnet,gff} and lightweight architectures for real-time performance~\cite{bfmnet,DDRNets}.
Transformer-based architectures~\cite{vaswani2017attention} further advanced segmentation~\cite{segformer,segmenter} by expanding perceptual fields, yielding greater accuracy. However, image-only methods often falter in adverse conditions, where domain adaptation~\cite{robust1,robust2} is used to transfer knowledge between weather domains. Integrating radar with camera data could address these limitations.

\subsubsection{Radar-based Semantic Segmentation}
Radar-only segmentation methods are classified into tensor-based and point-cloud-based approaches. Tensor-based methods~\cite{ouaknine2021multi,dalbah2024transradar} process and segment radar Range-Angle-Doppler (RAD) tensor, similar to the image segmentation task, whereas point cloud segmentation assigns labels to individual points. Existing algorithms~\cite{zeller2022gaussian,gan_radar,zeller2023radar_instance} focus on extracting reliable radar-specific features and mitigating the sparsity and noisy issues.



\subsubsection{Camera-Radar Fusion-based Semantic Segmentation}


Fusion-based methods integrate complementary information from radar and camera for segmentation tasks. CMGGAN~\cite{lekic2019automotive} fuses radar-generated images with RGB data, while SO-NET~\cite{john2020so} and MCAF-Net~\cite{multi} project radar points onto the image plane for detection and segmentation. BEV-based methods~\cite{schramm2024bevcar,radsegnet} jointly process radar and camera data. Achelous~\cite{guan2023achelous} integrates radar and monocular camera data for object detection and segmentation but primarily utilizes radar for detection, leaving image features for segmentation. Effectively leveraging radar point clouds for dense image-plane segmentation remains an unresolved challenge.

To the best of our knowledge, no existing research has yet explored how radar point clouds can support semantic segmentation in the image plane. Given the inherent sparsity of radar data and the lack of shape information in point cloud data, effectively leveraging radar point clouds for image-plane semantic segmentation remains an open challenge.

\subsection{Generative Methods in Segmentation}

Generative models are increasingly applied to improve semantic segmentation. GMMSeg~\cite{liang2022gmmseg} combines generative and discriminative approaches, leveraging Gaussian Mixture Models to enhance robustness to out-of-distribution data. FissGAN~\cite{fissgan} employs dual GANs to handle foggy conditions, generating edge information and semantic features.

Inpainting methods reconstruct missing image regions using segmentation masks. SGE-Net~\cite{liao2020guidance} uses segmentation maps for structural refinement, ensuring coherent inpainted results. Inpaint Anything~\cite{inpaintany} integrates Segment-Anything~\cite{sam} and Stable Diffusion~\cite{rombach2022high} to refine specific regions. While promising, these methods are yet to realize their full potential in enhancing semantic segmentation performance for camera-radar fusion.

\begin{figure*}[t]
\centering
\vspace{0.18cm}
\includegraphics[width=0.72\linewidth]{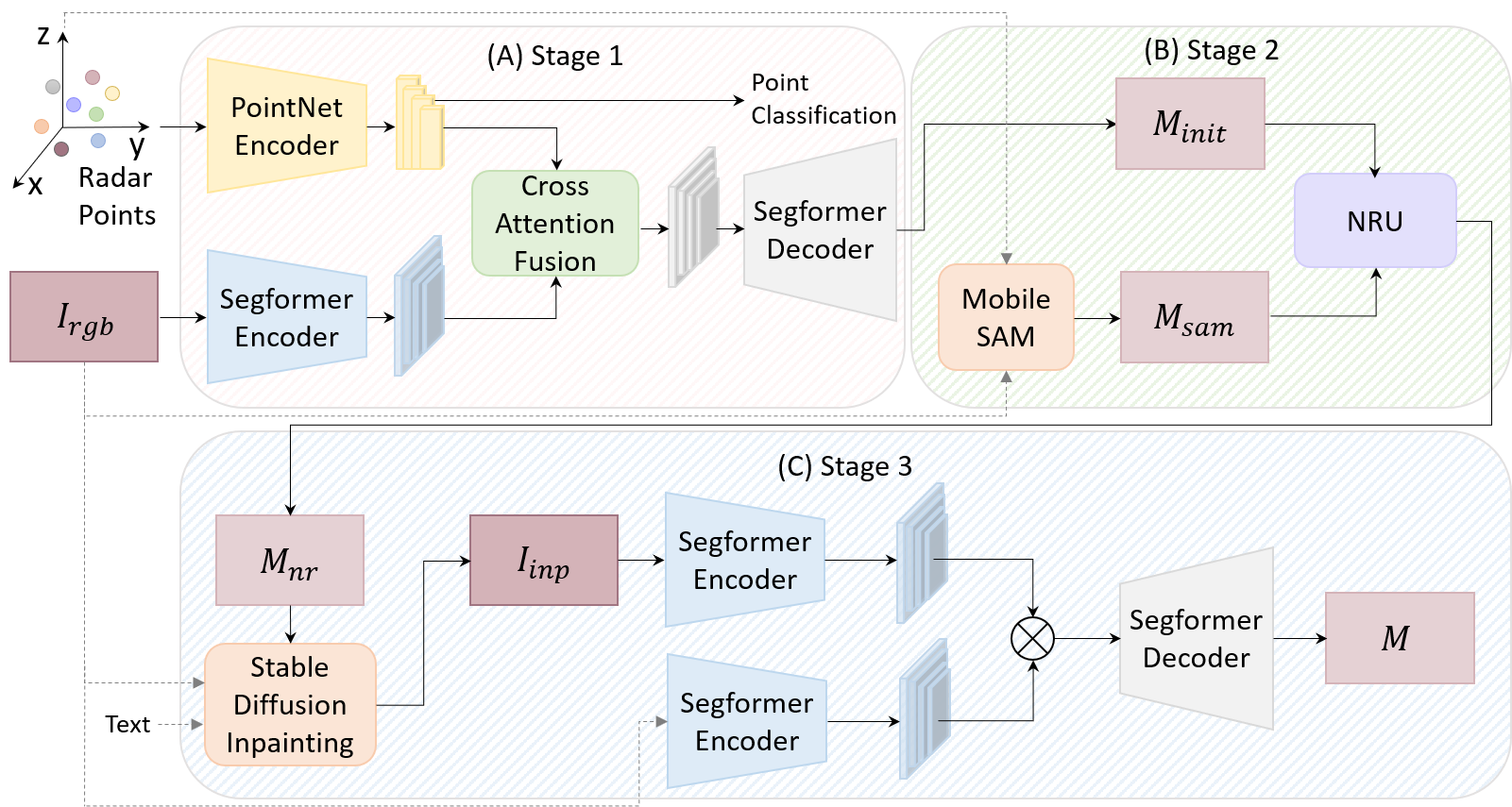}
   \caption{Model Architecture: Our three-stage framework combines radar and camera data for robust segmentation, especially in challenging conditions. First, radar and image features are extracted and fused through cross-attention to produce an initial segmentation mask. To improve resilience, MobileSAM generates an additional mask using radar points as prompts, and the Noise Reduction Unit (NRU) refines this by filtering radar-related noise. Finally, dual Segformer Encoders further enhance the refined mask, producing a high-quality segmentation output.}
\label{fig:model}
\vspace{-6mm}
\end{figure*}

\section{Approach}
This section begins by presenting the overall architecture of the proposed three-stage CaRaFFusion model. We then describe each stage in detail: the Camera-Radar Fusion stage, the Pseudo-Mask Generation and Mask Denoising stage, and the Image Inpainting and Final Mask Prediction stage. Lastly, we introduce the loss function used to train the networks.
\subsection{Model Architecture}
The proposed framework consists of three main stages, designed to process radar points and camera images to produce accurate segmented masks. 

In the first stage, an initial camera-radar fusion model is introduced for the semantic segmentation task. This model takes the radar point cloud and RGB image as inputs, which are processed by separate encoders. The extracted features are fused through a cross-attention strategy~\cite{vaswani2017attention} and passed to the Segformer decoder, producing initial segmentation masks $M_{\text{init}}$. Simultaneously, the radar features are processed by a PointNet~\cite{pointnet} decoder for point cloud classification.

The second stage focuses on generating more robust masks for the third stage. Here, MobileSAM~\cite{zhang2023mobilesam} is employed to generate additional SAM masks $M_{\text{sam}}$, using the image as input and radar points as prompts. This leverages the radar sensor's robustness to detect unseen objects, which may be obscured by blurriness or water droplets in the image. However, radar noise presents challenges, as object-bound radar points projected onto the image plane may lie on water surfaces, creating noisy SAM masks. To address this, we propose a Noise Reduction Unit (NRU) designed to filter out radar projection noise and refine the MobileSAM-generated masks, yielding $M_{\text{nr}}$ masks for the next stage.

The third stage takes the refined masks $M_{\text{nr}}$, the RGB image, and predicted object types from the radar point cloud classifier as inputs to generate an inpainted image. This process utilizes the power of a diffusion model to fill regions of the image blurred or obscured by water droplets. The inpainted image serves as an additional modality, processed by a separate Segformer encoder. Features encoded from both the original and inpainted images are concatenated and sent to the decoder, outputting the final segmentation masks.
\subsection{Stage 1: Camera-Radar Fusion}
This stage takes radar and camera data as inputs, utilizing the multi-task learning strategy to predict segmentation masks and classify the radar point cloud simultaneously. 
\noindent \textbf{Segmentation: }
The radar data, represented as points with shape $(N_{p}, 5)$, are processed through a PointNet~\cite{pointnet} encoder to extract spatial features. Here, $N_{p}$ represents the number of radar points, and $5$ corresponds to the five input features: the 3D positions of the points, Radar Cross Section, and Doppler velocities. The resulting radar features $F_{\text{radar}}^{i}$ have a shape of $(N, C_{r}^{i})$, where $N$ is the number of radar point features, and $C_{r}^{i}$ is the number of feature channels.

Simultaneously, the RGB image $I_{\text{rgb}}$ is processed by a Segformer encoder~\cite{segformer}, which extracts rich visual features $F_{\text{img}}^{i}$ with a shape of $(H^{i}, W^{i}, C^{i}_{I})$, where $H^{i}$ and $W^{i}$ denote the height and width of the image features, and $C^{i}_{I}$ is the number of image feature channels, with $C_{r}^{i} = C_{I}^{i}$. Here, $i \in \{1, 2, 3, 4\}$ refers to the layer index. To enhance the feature representations, we applied a cross-attention mechanism~\cite{vaswani2017attention} from the image features to the radar features. In our setup, we use the image features as query $Q_{\text{img}}$ and the radar features as the key $K_{\text{radar}}$ and value $V_{\text{radar}}$ matrices. The Cross-Attention Fusion (CAF) process is represented as follows: 
\begin{align}
    Q_{\text{img}}^{i}&=F_{\text{img}}^{i}W_{Q}^{i}, \\
    K_{\text{radar}}^{i}&=F_{\text{radar}}^{i}W_{K}^{i},\\
    V_{\text{radar}}^{i}&=F_{\text{radar}}^{i}W_{V}^{i},
\end{align}
\begin{equation}
\resizebox{0.43\textwidth}{!}{$
    \text{Attention}(Q_{\text{img}}^{i}, K_{\text{radar}}^{i}, V_{\text{radar}}^{i}) = \text{softmax} \left( \frac{Q_{\text{img}}^{i} (K_{\text{radar}}^{i})^T}{\sqrt{C^{i}_{I}}} \right) V_{\text{radar}}^{i},$}
\end{equation}
where $ Q_{\text{img}}^{i} \in \mathbb{R}^{C_{I}^{i} \times d} $ and $K_{\text{radar}}^{i}, V_{\text{radar}}^{i} \in \mathbb{R}^{C_{r}^{i} \times d}$. $Q_{\text{img}}^{i}$ are set to the extracted intermediate Segformer encoder features and $K_{\text{radar}}^{i}$, $V_{\text{radar}}^{i}$ are set to the extracted intermediate PointNet encoder features.
Then we get the final fused feature $F^i$: 
\begin{equation}
F^{i}=Q_{\text{img}}^{i} + \text{Attention}(Q_{\text{img}}^{i}, K_{\text{radar}}^{i}, V_{\text{radar}}^{i})
\end{equation}

The fused features $F^i, i\in \{1, 2, 3, 4\}$ generated by the CAF module are subsequently passed through a Segformer decoder, which produces the initial segmentation mask $M_{\text{init}}$. This mask integrates spatial information from radar data with visual context from the camera image, providing a solid foundation for further refinement in subsequent stages.

\noindent \textbf{Radar Point Cloud Segmentation:}
We train the radar point cloud segmentation task alongside the segmentation task for two main reasons. First, as shown in previous studies~\cite{multi,schramm2024bevcar}, a multitask learning strategy enables the model to extract more meaningful features, thus improving the performance of the primary task. Second, point cloud segmentation prepares for the second stage, where the MobileSAM model uses radar points as prompts to generate additional masks. Since MobileSAM is an instance segmentation model, it produces a binary mask for each prompt without identifying the object's category. By using the predicted class of a given radar point, we can assign a class label to each segmented binary mask.

The point cloud segmentation model functions as follows: encoded radar point features are first extracted by the PointNet encoder, which captures relevant spatial and contextual patterns. These features are then fed into a classification head consisting of multiple Multi-Layer Perceptrons (MLPs) followed by a softmax layer to classify the points.

\subsection{Stage 2: Pseudo-Mask Generation and Mask Denoising}
In stage 2, we improve $M_{\text{init}}$ by addressing limitations that arise under challenging conditions. Although radar data remain robust in adverse weather, they lack the fine detail needed to accurately outline object shapes, which is essential for segmentation. Consequently, despite the fusion of both modalities in stage 1, the predicted masks may still miss parts of objects. For example, in poor weather conditions, such as rain or water droplets in the lens, $M_{\text{init}}$ may struggle to maintain accuracy due to interference in visual data.

To supplement $M_{\text{init}}$, we employ MobileSAM~\cite{zhang2023mobilesam}, a lightweight Segment Anything Model, to generate additional masks $M_{\text{sam}}$ using radar points as prompts. This approach leverages the radar's robustness under adverse conditions, helping to produce masks that are less affected by weather-related interference. However, radar data can introduce noise, as radar points projected onto a 2D plane may inaccurately outline object boundaries, with some points positioned on water surfaces or other extraneous areas. This can create artifacts in $M_{\text{sam}}$, introducing excessive noise that impacts the third stage.

Since our first stage model segments the background $M_{\text{background}}$ and $M_{\text{water}}$, which are extracted from the first and last channel of $M_{\text{init}}$, with high accuracy, we can use $M_{\text{init}}$ to remove noisy artifacts from $M_{\text{sam}}$. To accomplish this, we introduce a simple Noise Reduction Unit (NRU). This module removes the background $M_{\text{background}}$ and waterline $M_{\text{water}}$ from $M_{\text{sam}}$ on a channel-wise basis, then adds the segmented object parts from $M_{\text{init}}$ back into $M_{\text{sam}}$ to create the final pseudo-mask $M_{\text{nr}}$. This refined mask $M_{\text{nr}}$ is used by the inpainting model in the next stage. The NRU process can be formally defined as shown in the following:
\begin{enumerate}
    \item \textbf{Step (1):} We define the noise mask \( M_{\text{noise}} \) as the sum of two components, \( M_{\text{background}} \) and \( M_{\text{water}} \):
    \[
    M_{\text{noise}} := M_{\text{background}} + M_{\text{water}}.
    \]
    This step identifies the noise sources within the mask by combining background and water areas, which are often sources of irrelevant or misleading information in segmentation.

    \item \textbf{Step (2):} The denoised mask \( M_{\text{denoised}} \) is computed by subtracting the noise mask \( M_{\text{noise}} \) from the initial mask \( M_{\text{sam}} \) generated by MobileSAM:
    \[
    M_{\text{denoised}} := M_{\text{sam}} - M_{\text{noise}}.
    \]
    This operation removes unwanted background and water information, retaining only relevant regions that better represent the object.

    \item \textbf{Step (3):} A ReLU activation function is applied to \( M_{\text{denoised}} \), resulting in the final denoised mask \( M_{\text{sam}} \):
    \[
    M_{\text{sam}} := \text{ReLU}(M_{\text{denoised}}).
    \]
    ReLU effectively removes any negative values that may have appeared from the subtraction, ensuring that only non-negative values are kept.

    \item \textbf{Step (4):} The merged mask \( M_{\text{merged}} \) is then created by combining the refined mask \( M_{\text{sam}} \) with the initial segmentation mask \( M_{\text{init}} \) from Stage 1:
    \[
    M_{\text{merged}} := M_{\text{sam}} + M_{\text{init}}.
    \]
    This merging step leverages the strengths of both the refined MobileSAM output and the initial mask to create a more comprehensive representation.

    \item \textbf{Step (5):} Finally, the noise-reduced mask \( M_{\text{nr}} \) is obtained by applying a clamping operation to \( M_{\text{merged}} \), restricting its values within the range \([0, 1]\):
    \[
    M_{\text{nr}} := \text{clamp}(M_{\text{merged}}, 0, 1).
    \]
    This step ensures that the mask values remain within valid bounds, enhancing stability of the final output.
\end{enumerate}
Each of these steps is applied channel-wise, allowing the model to handle multi-channel data independently for more precise noise reduction and mask refinement.

\subsection{Stage 3: Image Inpainting and Final Mask Prediction}
The final stage aims to generate robust segmentation masks by leveraging the capabilities of a zero-shot generative model. First, the Stable Diffusion model~\cite{rombach2022high} takes the mask output $M_{\text{nr}}$ from stage 2, the related text prompt of the mask and the original input image $I_{\text{img}}$ as inputs, producing an inpainted image $I_{\text{inp}}$ of the same size as the original image. This process serves multiple purposes: during adverse weather conditions, such as rain, images often become blurry and are affected by water droplets. The robust masks $M_{\text{nr}}$ generated in stage 2 mitigate these effects by focusing on reliable object regions, allowing the inpainting model to generate plausible object details in the masked areas. The inpainting process is summarized in Algorithm \ref{alg:inpainting}.

Subsequently, the inpainted image $I_{\text{inp}}$ and the original image $I_{\text{img}}$ are processed separately by two distinct Segformer encoders. This parallel encoding enables the model to capture information from both the original image and the enhanced inpainted image, which is especially beneficial under challenging weather conditions. The outputs from these encoders are concatenated to integrate spatial and semantic features from both sources and then fed into a Segformer decoder, which generates the final segmentation mask $M$ for each class. 

This final decoding step synthesizes the combined features, producing an accurate and consistent mask with reduced noise and enhanced detail. By leveraging multi-modal refinement, where the inpainted and original images complement each other, the model achieves a high-quality segmentation output even in adverse conditions.
\vspace{-2mm}
\begin{algorithm}
\caption{Iterative Inpainting}
\label{alg:inpainting}
\begin{algorithmic}[1]
\Require Image $I_{\text{img}}$, Set of text prompts $\{P_1, P_2, \dots, P_N\}$, Set of masks $\{M_1, M_2, \dots, M_N\}$, Stable Diffusion model $\text{SD}$
\State Initialize inpainted image $I \gets I_{\text{img}}$
\For{each mask $M_i$ in $\{M_1, M_2, \dots, M_N\}$}
    \State $I \gets \text{SD}(I, M_i, P_i)$ 
\EndFor
\State \Return $I$ 

\end{algorithmic}
\end{algorithm}
\vspace{-5mm}
\subsection{Loss Functions}
Two loss functions are used to train our network. For radar point cloud segmentation in stage 1, we apply Focal Loss~\cite{Lin2017FocalLF}, defined as follows:
\begin{equation}
\vspace{-2mm}
    \label{eq:focal}   
L_{\text{cls}} = - \sum_{c=1}^{C} \alpha_{c}(1 - p_{c})^{\gamma} \log(p_{c}),
\end{equation}
where \( p_{c} \) represents the predicted probability of class $c$, and \( C \) denotes the number of classes. We choose the parameters $\alpha_c$ according to the relative class frequency of class $c$ and set $\gamma=2$ to balance class distribution and emphasize harder samples over easier ones, respectively\cite{Lin2017FocalLF}.

For 2D semantic segmentation, we use the Dice loss~\cite{Sudre2017dice}, defined as follows:
\begin{equation}
    \label{eq:dice}
L_{\text{seg}} = 1 - \frac{2 \sum_{i=1}^{N} p_i g_i}{\sum_{i=1}^{N} p_i + \sum_{i=1}^{N} g_i},
\end{equation}
where \( p_{i} \) and \( g_{i} \) denote the predicted probability and the ground truth label for the \( i^{th} \) pixel, respectively, and \( N \) is the total number of pixels in the image. The Dice loss is particularly effective for addressing class imbalance, as it emphasizes the overlap between the predicted and ground truth segmentation masks.

\section{Experiments}

This section begins with an explanation of the dataset and implementation details. Next, we evaluate our results both quantitatively and qualitatively. Finally, we perform ablation studies to further demonstrate the effectiveness of the proposed methods.
\subsection{Dataset and the Implementation Details}

\noindent \textbf{Dataset Details: }
We train and test our method on the Waterscenes dataset~\cite{yao2024waterscenes}, which provides per-pixel segmentation masks for seven object classes, as well as a waterline class. Additionally, we include a background class to aid in mask denoising for the Noise Reduction Unit (NRU). Furthermore, we evaluate our method on a subset of the dataset, selecting images based on specific weather conditions to test performance under the most challenging scenarios, such as  \emph{foggy}, \emph{strong light exposure}, and \emph{rainy} conditions, as well as instances where the camera sensor is distorted, e.g., by \emph{water drop hit}. The dataset comprises a total of 37,884 training, 10,824 validation, and 1,596 test images.

\noindent \textbf{Implementation Details: }
We use the Segformer-B0 architecture for both the encoder and decoder and the Stable Diffusion Inpainting models from the Huggingface platform~\cite{wolf2019huggingface}. During training, if the number of radar points exceeds 1000, we randomly sample 1000 points; otherwise, we apply zero-padding to the input. The images are downscaled to 320x320. For efficient and stable training, we utilize the PyTorch Lightning library~\cite{falcon2019pytorch}. Our models are trained on an Nvidia\textsuperscript{\textregistered} Tesla A30 GPU with a batch size of 32.

To optimize computational efficiency, we first train the stage 1 model and use the NRU to generate masks. Stage 3 is then trained separately using these extracted masks. We employ the AdamW optimizer~\cite{loshchilov2017decoupled} with an initial learning rate of $5 \times 10^{-4}$, which is linearly decayed to a minimum learning rate of $1 \times 10^{-6}$. In line with the training strategy from Achelous~\cite{guan2023achelous}, no additional data augmentation techniques are applied during training.

\subsection{Quantitative Results}
\vspace{-2mm}
\begin{table}[htp]
    \centering
    \small
    \resizebox{8cm}{!} 
{
    \begin{tabular}{l||ccc}
        \hline
        \textbf{Method} & \textbf{mIoU (\%)} & Params (M)  \\ \hline

        Achelous-EV-GDF-PN-S2~\cite{guan2023achelous} & - & 8.28 
        \\ \hline
        Segformer(camera-only) & 75.47 & 3.7
        \\ 
        Segformer Fusion & 76.62  & 4.3
        \\ 
        Segformer Fusion + Inpainting & \textbf{78.10} &  4.3 + 9.66(MobileSAM) + 1290(Stable Diffusion)
        \\ \hline
    \end{tabular}
    }
    \caption{Comparison of Segmentation Methods with mIoU under Adverse Weather Conditions.}
    \label{table:results}
\vspace{-3mm}
\end{table}

To evaluate our method, we use the mean Intersection over Union (mIoU) metric~\cite{pascal}, a widely recognized measure for assessing segmentation quality. 
We first evaluate the models on the adverse weather sub-testset, with results presented in Table \ref{table:results}. There is a clear improvement in mIoU from the camera-only model to the fusion model, and further to the fusion model with inpainting. This trend indicates that incorporating multiple modalities enhances segmentation accuracy by providing richer information about the scene, especially during the hard sample segmentation scenarios. Additionally, inpainting contributes to improved results by filling in missing or obstructed regions, especially in cases where visual data is compromised due to adverse weather conditions. It is important to note that the Achelous models did not include evaluation results on the adverse weather subset. Therefore, our comparison with the best-performing model from~\cite{guan2023achelous} is limited to model parameter counts only.

Additionally, we analyze the models on the total test set, which contains 5120 images, against the models provided in the Achelous~\cite{guan2023achelous} as shown in the Table \ref{table:results_total}. Our fusion-based methods for semantic segmentation of targets achieve 8.11\% improvement compared to the best Achelous model. However, due to the inherent noise in radar data, some point segmentations are incorrectly predicted, for example, due to water reflections. This results in a slight performance drop in $\text{mIoU}_d$ when compared to the Achelous models.

\begin{table}[htp]
    \centering
    \small
    \caption{Comparison of Segmentation Methods with mIoU on the Total Testset.}
    \label{table:results_total}
    \resizebox{8cm}{!} 
{
    \begin{tabular}{l||cc}
        \hline
        \textbf{Method} & $^{\star}$$\text{mIoU}_{t}$ (\%) & $^{\star}$$\text{mIoU}_{d}$ (\%) \\ \hline
        Achelous-MV-GDF-PN-S0$^{\dag}$~\cite{guan2023achelous} & 70.60 & 99.5 \\
        Achelous-MV-GDF-PN-S1$^{\dag}$~\cite{guan2023achelous} & 73.20 & 99.5 \\
         Achelous-EV-GDF-PN-S2$^{\dag}$~\cite{guan2023achelous} & 74.10 & 99.5 
        \\ \hline
        Segformer(camera-only) & 81.12 & 98.64 
        \\ 
        Segformer Fusion & 82.39 & 98.74 
        \\
        Segformer Fusion + Inpainting & 82.21 & 98.75  
        \\ \hline
    \end{tabular}
    }
    \begin{tablenotes}
    \small
      \item $^{\star}$$\text{mIoU}_{t}$: mIoU of targets, $^{\star}$$\text{mIoU}_{d}$: mIoU of drivable area 
      \small
      \item $^{\dag}$ These results come from the original paper. 
    \end{tablenotes}
      \vspace{-4mm}
\end{table}
Here, we use `Segformer Fusion' and `Segformer Fusion + Inpainting' to denote the models from the first and third stages, respectively. 
\newcommand{\subfigurewidth}{0.16\textwidth}
\begin{figure*}[ht]
    \centering
    \begin{subfigure}[t]{\subfigurewidth}
        \centering
        \caption{Input Img}
        \includegraphics[width=\textwidth]{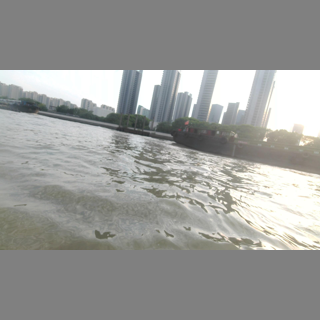} 
        \label{fig:subfig1}
    \end{subfigure}
    \begin{subfigure}[t]{\subfigurewidth}
        \centering
        \caption{Inpaint Img}
        \includegraphics[width=\textwidth]{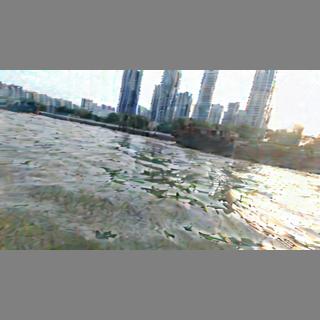} 
        \label{fig:subfig1}
    \end{subfigure}
    \begin{subfigure}[t]{\subfigurewidth}
        \centering
        \caption{GT}
        \includegraphics[width=\textwidth]{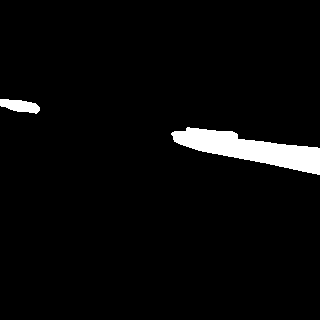} 
        \label{fig:subfig2}
    \end{subfigure}
    \begin{subfigure}[t]{\subfigurewidth}
        \centering
        \caption{Baseline}
        \includegraphics[width=\textwidth]{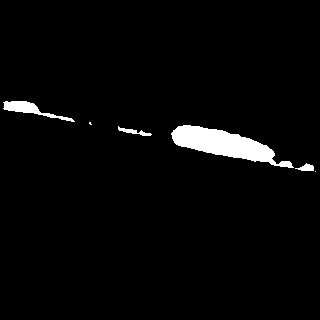} 
        \label{fig:subfig3}
    \end{subfigure}
    \begin{subfigure}[t]{\subfigurewidth}
        \centering
        \caption{Fusion}
        \includegraphics[width=\textwidth]{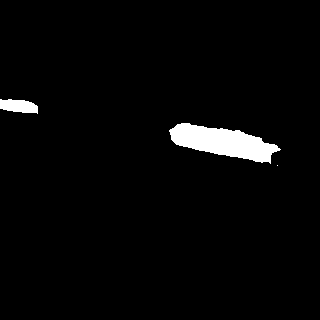} 
        \label{fig:subfig4}
    \end{subfigure}
    \begin{subfigure}[t]{\subfigurewidth}
        \centering
        \caption{Ours}
        \includegraphics[width=\textwidth]{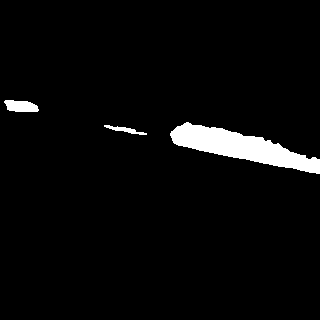} 
        \label{fig:subfig5}
    \end{subfigure}
\vspace{-3mm}
    
    \begin{subfigure}[t]{\subfigurewidth}
        \centering
        \includegraphics[width=\textwidth]{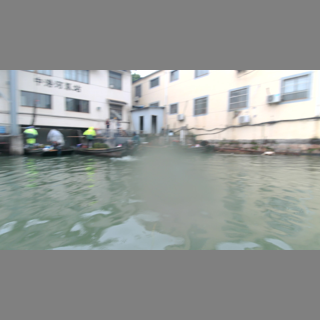} 
        \label{fig:subfig1}
    \end{subfigure}
    \begin{subfigure}[t]{\subfigurewidth}
        \centering
        \includegraphics[width=\textwidth]{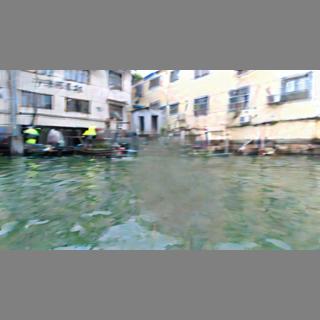} 
        \label{fig:subfig1}
    \end{subfigure}
    \begin{subfigure}[t]{\subfigurewidth}
        \centering
        \includegraphics[width=\textwidth]{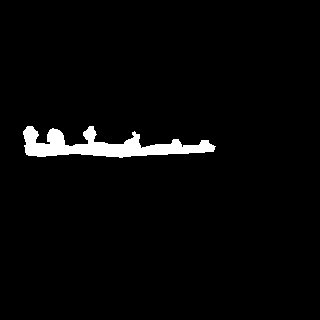} 
        \label{fig:subfig2}
    \end{subfigure}
    \begin{subfigure}[t]{\subfigurewidth}
        \centering
        \includegraphics[width=\textwidth]{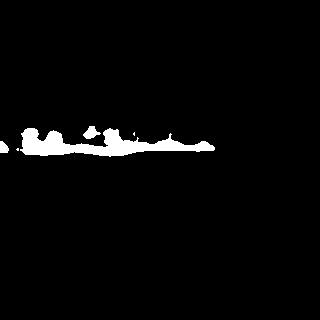} 
        \label{fig:subfig3}
    \end{subfigure}
    \begin{subfigure}[t]{\subfigurewidth}
        \centering
        \includegraphics[width=\textwidth]{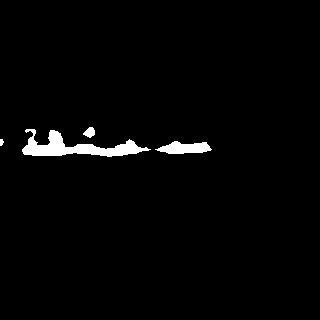} 
        \label{fig:subfig4}
    \end{subfigure}
    \begin{subfigure}[t]{\subfigurewidth}
        \centering
        \includegraphics[width=\textwidth]{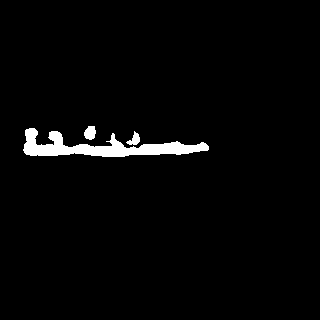} 
        \label{fig:subfig5}
    \end{subfigure}
\vspace{-3mm}

    \begin{subfigure}[t]{\subfigurewidth}
        \centering
        \includegraphics[width=\textwidth]{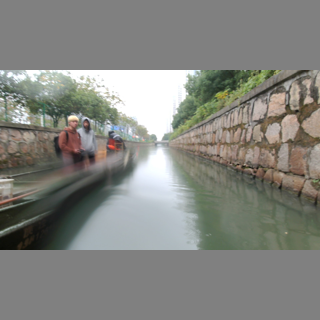} 
        \label{fig:subfig1}
    \end{subfigure}
    \begin{subfigure}[t]{\subfigurewidth}
        \centering
        \includegraphics[width=\textwidth]{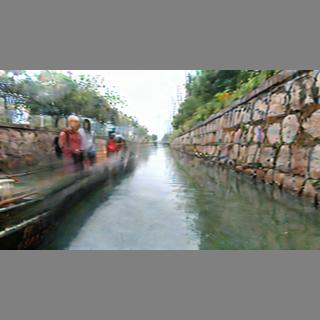} 
        \label{fig:subfig1}
    \end{subfigure}
    \begin{subfigure}[t]{\subfigurewidth}
        \centering
        \includegraphics[width=\textwidth]{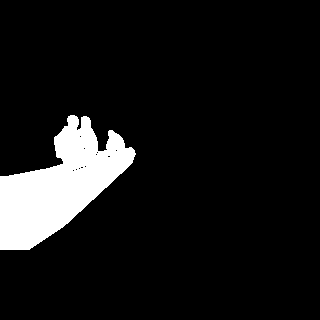} 
        \label{fig:subfig2}
    \end{subfigure}
    \begin{subfigure}[t]{\subfigurewidth}
        \centering
        \includegraphics[width=\textwidth]{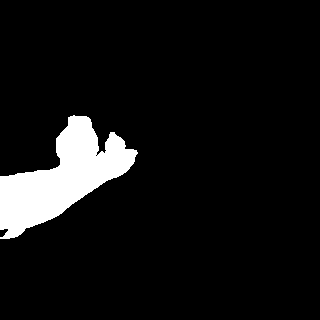} 
        \label{fig:subfig3}
    \end{subfigure}
    \begin{subfigure}[t]{\subfigurewidth}
        \centering
        \includegraphics[width=\textwidth]{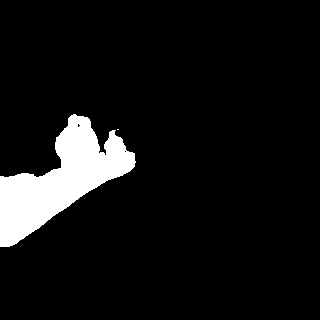} 
        \label{fig:subfig4}
    \end{subfigure}
    \begin{subfigure}[t]{\subfigurewidth}
        \centering
        \includegraphics[width=\textwidth]{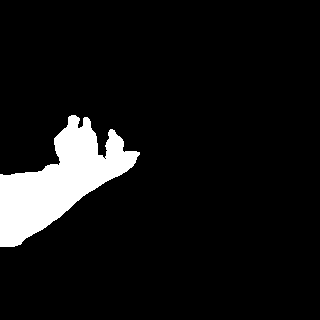} 
        \label{fig:subfig5}
    \end{subfigure}

    \vspace{-3mm}

    \begin{subfigure}[t]{\subfigurewidth}
        \centering
        \includegraphics[width=\textwidth]{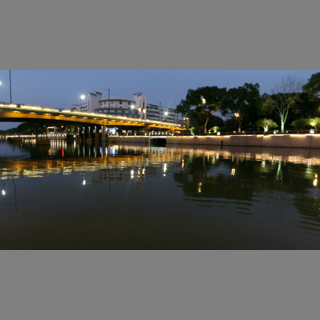} 
        \label{fig:subfig1}
    \end{subfigure}
    \begin{subfigure}[t]{\subfigurewidth}
        \centering
        \includegraphics[width=\textwidth]{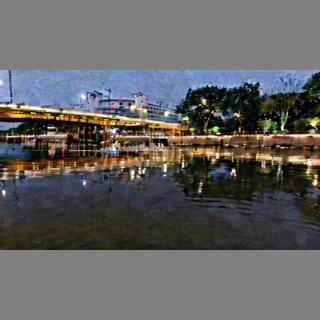} 
        \label{fig:subfig1}
    \end{subfigure}
    \begin{subfigure}[t]{\subfigurewidth}
        \centering
        \includegraphics[width=\textwidth]{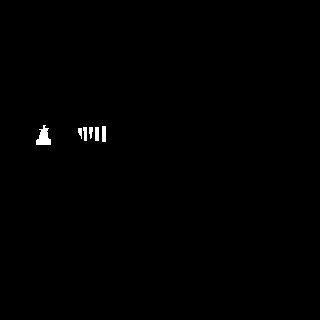} 
        \label{fig:subfig2}
    \end{subfigure}
    \begin{subfigure}[t]{\subfigurewidth}
        \centering
        \includegraphics[width=\textwidth]{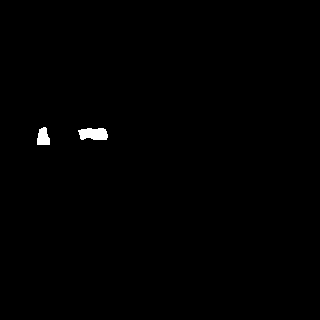} 
        \label{fig:subfig3}
    \end{subfigure}
    \begin{subfigure}[t]{\subfigurewidth}
        \centering
        \includegraphics[width=\textwidth]{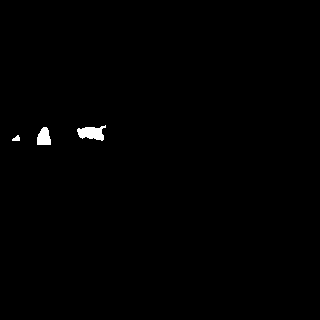} 
        \label{fig:subfig4}
    \end{subfigure}
    \begin{subfigure}[t]{\subfigurewidth}
        \centering
        \includegraphics[width=\textwidth]{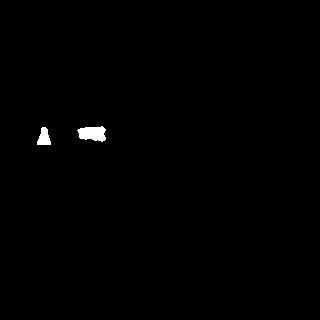} 
        \label{fig:subfig5}
    \end{subfigure}

    \vspace{-3mm}

    \begin{subfigure}[t]{\subfigurewidth}
        \centering
        \includegraphics[width=\textwidth]{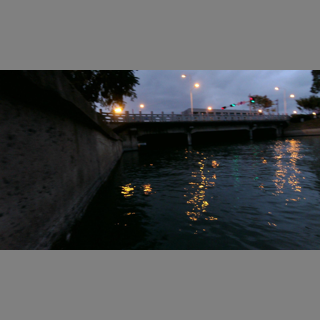} 
        \label{fig:subfig1}
    \end{subfigure}
    \begin{subfigure}[t]{\subfigurewidth}
        \centering
        \includegraphics[width=\textwidth]{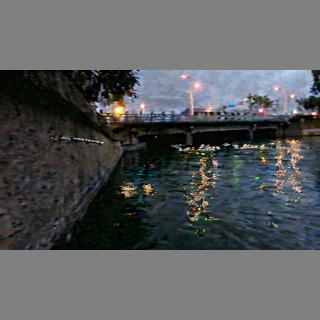} 
        \label{fig:subfig1}
    \end{subfigure}
    \begin{subfigure}[t]{\subfigurewidth}
        \centering
        \includegraphics[width=\textwidth]{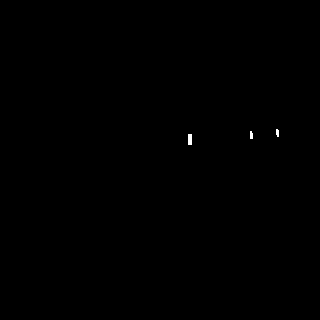} 
        \label{fig:subfig2}
    \end{subfigure}
    \begin{subfigure}[t]{\subfigurewidth}
        \centering
        \includegraphics[width=\textwidth]{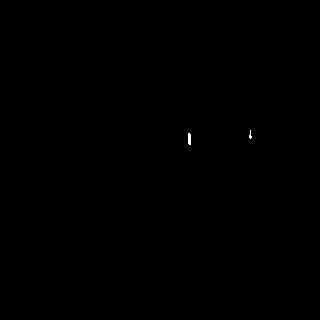} 
        \label{fig:subfig3}
    \end{subfigure}
    \begin{subfigure}[t]{\subfigurewidth}
        \centering
        \includegraphics[width=\textwidth]{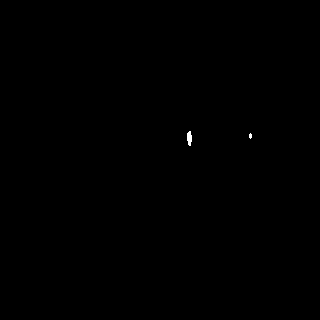} 
        \label{fig:subfig4}
    \end{subfigure}
    \begin{subfigure}[t]{\subfigurewidth}
        \centering
        \includegraphics[width=\textwidth]{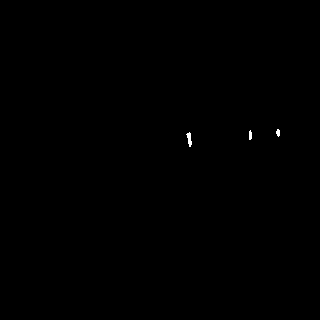} 
        \label{fig:subfig5}
    \end{subfigure}

    \vspace{-3mm}

    \begin{subfigure}[t]{\subfigurewidth}
        \centering
        \includegraphics[width=\textwidth]{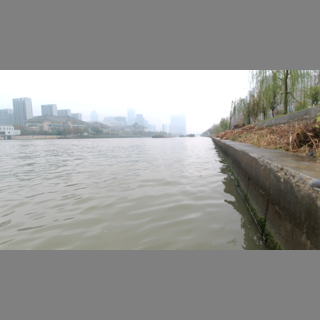} 
        \label{fig:subfig1}
    \end{subfigure}
    \begin{subfigure}[t]{\subfigurewidth}
        \centering
        \includegraphics[width=\textwidth]{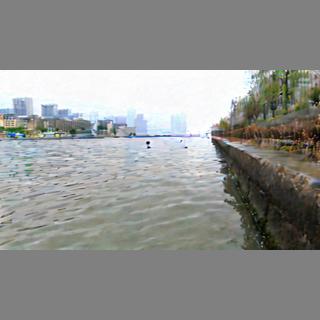} 
        \label{fig:subfig1}
    \end{subfigure}
    \begin{subfigure}[t]{\subfigurewidth}
        \centering
        \includegraphics[width=\textwidth]{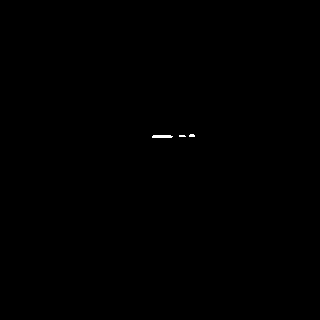} 
        \label{fig:subfig2}
    \end{subfigure}
    \begin{subfigure}[t]{\subfigurewidth}
        \centering
        \includegraphics[width=\textwidth]{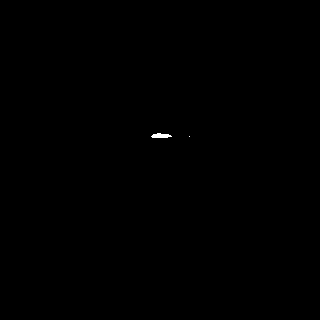} 
        \label{fig:subfig3}
    \end{subfigure}
    \begin{subfigure}[t]{\subfigurewidth}
        \centering
        \includegraphics[width=\textwidth]{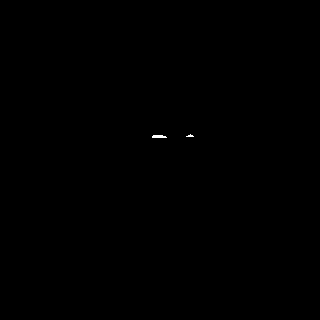} 
        \label{fig:subfig4}
    \end{subfigure}
    \begin{subfigure}[t]{\subfigurewidth}
        \centering
        \includegraphics[width=\textwidth]{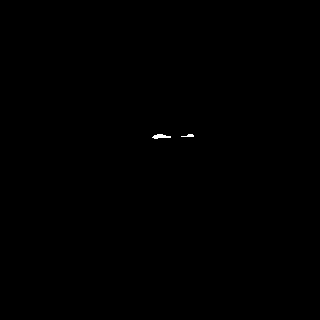} 
        \label{fig:subfig5}
    \end{subfigure}
\vspace{-4mm}

    \caption{Qualitative results show that our inpainting technique is likely addressing missing or occluded regions in the data. It helps to fill in parts of the objects or scenes that might otherwise go undetected, thereby boosting the model’s ability to achieve higher segmentation accuracy. Columns (a), (b), and (c) visualize the original image, the inpainted image, and the Ground Truth (GT) mask of the given image. Columns (d), (e), and (f) illustrate the segmentation from the image-only baseline, the fusion model from the first stage, and the final model from the third stage, respectively.}
    \label{fig:qualitative}
    \vspace{-6mm}
\end{figure*}

\subsection{Qualitative Results}
\noindent \textbf{Inpainted Images: }
The generated inpainted images exhibit a pixel distribution closely matching the original images, with the primary differences occurring in the inpainted areas. These regions, however, do not seamlessly integrate the original texture of the masked areas, which can result in a noticeable difference in appearance. Additionally, the inpainting process introduces a degree of noise, attributed to the inherent variability of the sampling process. Despite these challenges, such as the distinct textures and noise, the system demonstrates a remarkable ability to learn and fuse features from both the original and inpainted images effectively. This suggests that the model is robust in aligning and integrating visual information, enabling it to synthesize cohesive images even when texture inconsistencies are present. To demonstrate qualitatively that our method is effective in adverse conditions, some predicted images can be found in the second column of Figure \ref{fig:qualitative}.

\noindent \textbf{Segmentation: }
Building on previous analyses, our model demonstrates notable performance improvements in semantic segmentation tasks, as illustrated in Figure \ref{fig:qualitative}. We present qualitative comparisons between the baseline prediction, the fusion prediction from the first stage, and the final prediction incorporating inpainted images. The results show that the proposed CaRaFFusion model detects finer details, producing more accurate masks.
Despite challenges such as texture inconsistencies and noise in inpainted regions, the model effectively learns to integrate features from both original and inpainted areas. This capacity to reconcile pixel distribution differences and fuse diverse textures enables the generation of more precise segmentations. Consequently, these improvements yield a significant boost over the baseline, highlighting the model’s effectiveness in enhancing semantic segmentation performance.
\subsection{Ablation Studies}

This section first evaluates the sampling strategy for the radar point input. Then, we also evaluate the fusion approach at the third stage. 

\noindent \textbf{Input Point Sampling: }
Given that the number of radar points varies between frames, it is necessary to sample a subset of radar points during training. As shown in Table \ref{table:sample}, increasing the number of sampled radar points improves the effective utilization of radar point cloud data, leading to enhanced performance.
\begin{table}[H]
\vspace{-2mm}
    \centering
    \small
    \begin{tabular}{c||c}
        \hline Number of sampled points during training & mIoU(\%) \\ \hline
        100 & 75.10\\\hline
        200 & 75.39\\\hline
        1000 & 76.72\\\hline
    \end{tabular}
    \caption{Radar Points Sampling Comparison.}
    \label{table:sample}
    \vspace{-4mm}
\end{table}

\noindent \textbf{Different Fusion Methods: }
We evaluate the effectiveness of different image fusion methods in the third stage. Different fusion methods were implemented to assess their impact on segmentation performance. Table \ref{table:fusion} summarizes the performance of each fusion method.

\begin{table}[H]
\vspace{-1mm}
    \centering
    \small
    \begin{tabular}{c||c}
        \hline Fusion Method & mIoU(\%) \\ \hline
        Addition & 77.13\\\hline
        Gated Fusion~\cite{depth1} & 77.43\\\hline
        Concatenation & \textbf{78.10}\\\hline
    \end{tabular}
    \caption{Results of Different Fusion Methods in Stage 3.}
    \vspace{-3mm}
    \label{table:fusion}
\end{table}

\noindent \textbf{Importance of Inpainting Fusion:}
To demonstrate the effectiveness of image fusion in the third stage, we trained a model without inpainting fusion by directly using the inpainted images as input and evaluated it on the same test set. Table \ref{table:fusion_inpainting} shows the performance with and without fusion in the third stage. This indicates the importance of the fusion of original and inpainted images.

\begin{table}[H]
\vspace{-1mm}
    \centering
    \small
    \begin{tabular}{c||c}
        \hline  & mIoU(\%)\\ \hline
        Fusion & 78.10\\\hline No Fusion & 65.13 \\ \hline
    \end{tabular}
    \caption{Results with and without inpainting fusion in Stage 3.}
    \vspace{-3mm}
    \label{table:fusion_inpainting}
\end{table}
\section{Discussion and Limitations}
CaRaffusion is designed to operate with high-precision 3D radar point clouds that include height values. The rationale behind this choice is that accurate 3D positions are essential for correctly deriving 2D pixel coordinates when projecting 3D points onto the 2D image plane. This, in turn, enables the MobileSAM module to effectively utilize this information to generate inpainted images. Consequently, WaterScenes~\cite{yao2024waterscenes} is used for experiments instead of datasets like nuScenes~\cite{nuscenes}, where radar point clouds lack height information.

Integrating an image inpainting framework into our 2D semantic segmentation architecture creates an efficiency bottleneck. Advancements in diffusion model inference could resolve real-time constraints and we leave this as future work. In our work, we rather focused on a general concept, how to incorporate a diffusion model into a camera-radar fusion pipeline.

Radar data, while invaluable in adverse weather conditions, often contain significant noise that can degrade perception performance. Sources of noise include reflections from water surfaces, environmental clutter, and overlapping object signals. Future work could explore more advanced noise reduction techniques to generate more noise-free pseudo-masks for a better inpainting framework.

\section{Conclusion}
In this work, we presented a novel three-stage framework, CaRaFFusion, that effectively integrates radar and camera data for robust segmentation by integrating a generative image inpainting model, particularly suited for adverse weather conditions where traditional segmentation methods may struggle. By combining radar and camera data, our model leverages the unique strengths of each modality to compensate for visual limitations and environmental noise, which are prevalent challenges in outdoor settings. The framework approximately recovers essential shape features by fusing radar and image data and implementing the inpainting pipeline to enhance 2D pixel-wise semantic segmentation performance in adverse weather conditions. While our results show promise, future work should explore optimizing the framework’s efficiency and reducing GPU capacity requirements to enhance scalability and make it more feasible for real-time applications. Addressing these performance concerns will be crucial for deploying this model in resource-constrained environments.
\vspace{-1mm}
\section{Acknowledgement}
Research leading to these results 
has received funding from the EU ECSEL Joint Undertaking under grant agreement n° 101007326 (project AI4CSM) and from the partner national funding authorities the German Ministry of Education and Research (BMBF).

\bibliographystyle{IEEEtran}
\bibliography{main}


\end{document}